\documentclass[letterpaper, 10 pt, conference]{ieeeconf}  

\IEEEoverridecommandlockouts                              

\usepackage{graphics} 
\usepackage{graphicx}
\usepackage{epsfig} 
\usepackage{mathptmx} 
\usepackage{times} 
\usepackage{amsmath} 
\usepackage{amssymb}  
\usepackage{algorithm}
\usepackage{algpseudocode}
\usepackage[capitalize]{cleveref} 
\usepackage{subcaption}
\usepackage{booktabs}
\usepackage{array}
\usepackage{tikz}
\usetikzlibrary{positioning}

\title{\LARGE \bf
CRED: Counterfactual Reasoning and Environment Design \\for Active Preference Learning
}

\begin{document}

\author{Yi-Shiuan Tung, Gyanig Kumar, Wei Jiang, Bradley Hayes, Alessandro Roncone\\
Department of Computer Science, University of Colorado Boulder\\ \{\tt\small yi-shiuan.tung, gyanig.kumar, wei.jiang, bradley.hayes, alessandro.roncone\}@colorado.edu}


\maketitle
\thispagestyle{empty}
\pagestyle{empty}

\begin{abstract}
As a robot's operational environment and tasks to perform within it grow in complexity, the explicit specification and balancing of optimization objectives to achieve a preferred behavior profile moves increasingly farther out of reach. These systems benefit strongly by being able to align their behavior to reflect human preferences and respond to corrections, but manually encoding this feedback is infeasible.
Active preference learning (APL) learns human reward functions by presenting trajectories for ranking. However, existing methods sample from fixed trajectory sets or replay buffers that limit query diversity and often fail to identify informative comparisons. We propose CRED, a novel trajectory generation method for APL that improves reward inference by jointly optimizing environment design and trajectory selection to efficiently query and extract preferences from users. CRED ``imagines'' new scenarios through environment design and leverages counterfactual reasoning--by sampling possible rewards from its current belief and asking ``What if this were the true preference?''--to generate trajectory pairs that expose differences between competing reward functions. Comprehensive experiments and a user study show that CRED significantly outperforms state-of-the-art methods in reward accuracy and sample efficiency and receives higher user ratings.

\end{abstract}

\begin{figure}[t]
    \centering
    \begin{subfigure}{0.49\columnwidth}
        \includegraphics[width=\linewidth]{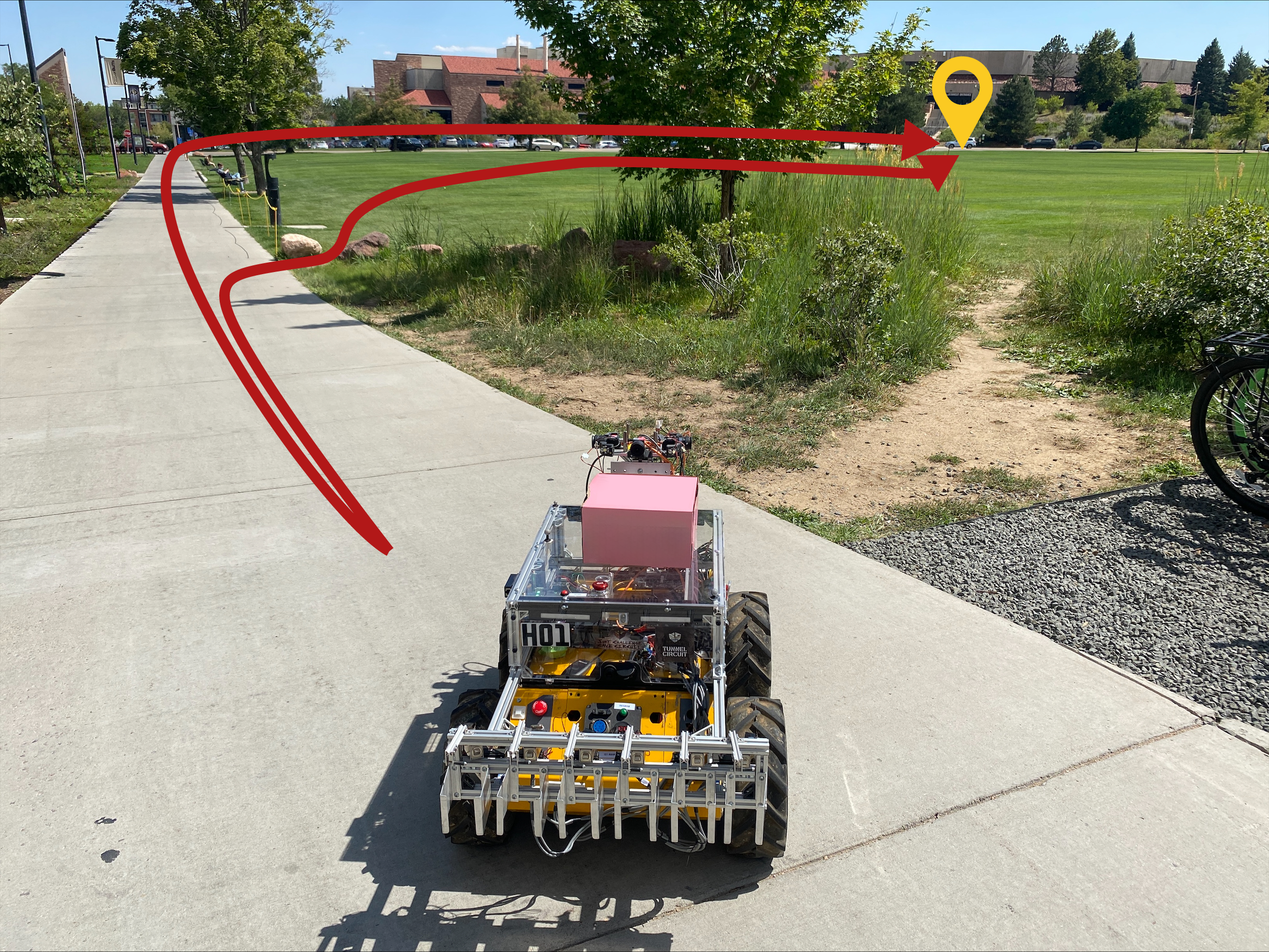} 
        \caption{Suboptimal queries}
    \end{subfigure}\hfill
    \begin{subfigure}{0.49\columnwidth}
        \includegraphics[width=\linewidth]{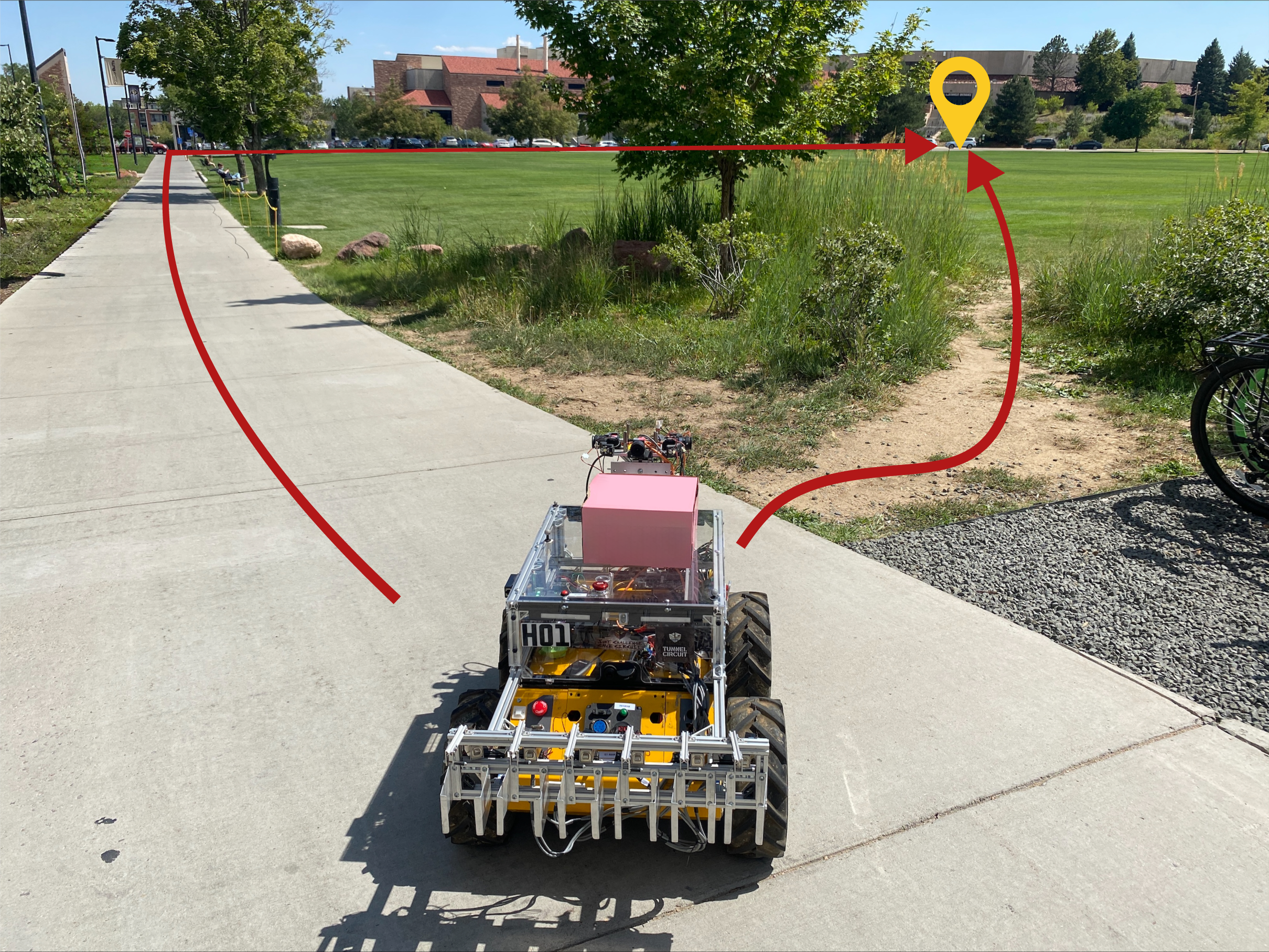} 
        \caption{Counterfactuals}
    \end{subfigure}\hfill
    \begin{subfigure}{0.49\columnwidth}
        \includegraphics[width=\linewidth]{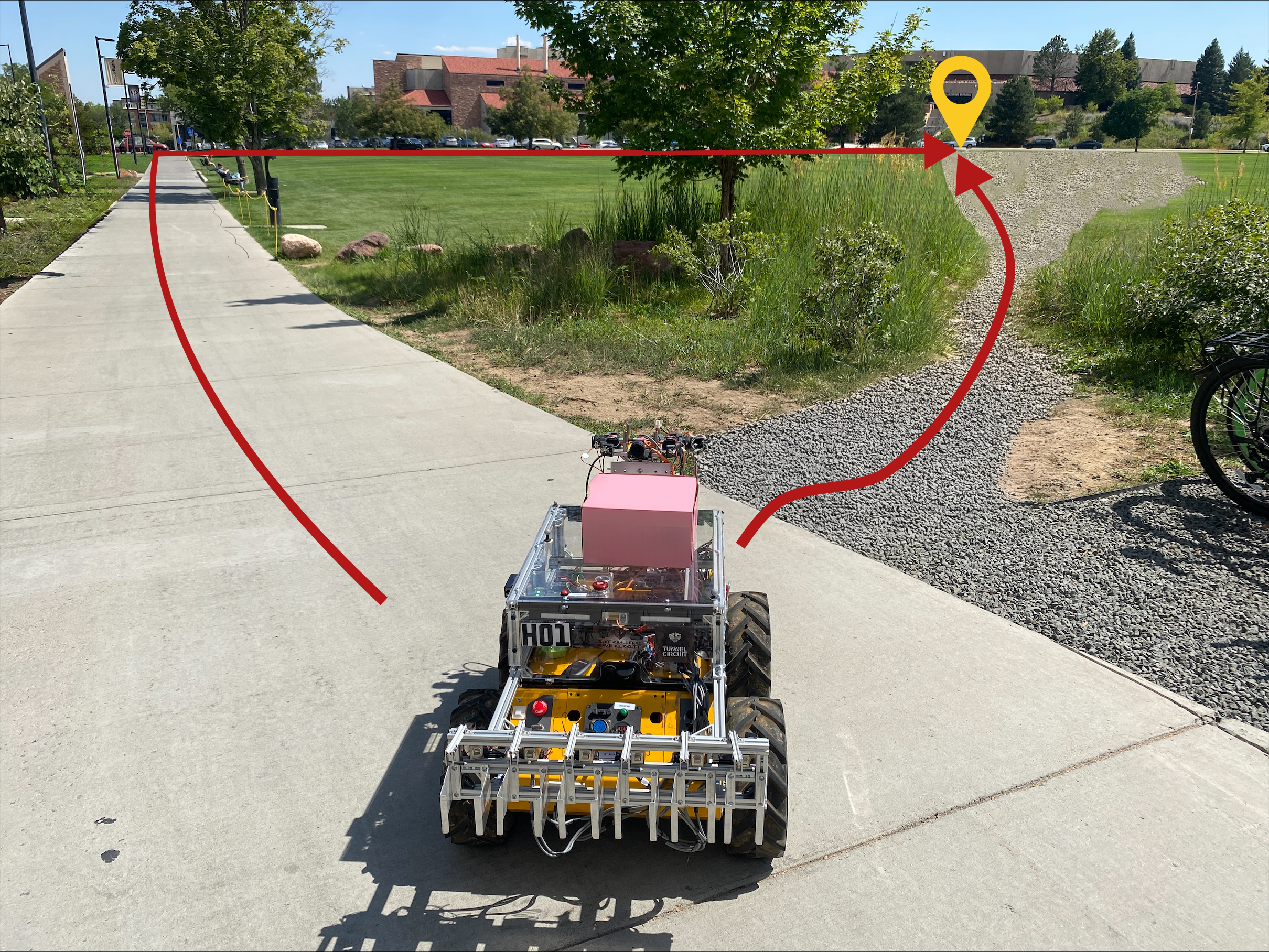} 
        \caption{Environment design}
    \end{subfigure}\hfill
    \begin{subfigure}{0.49\columnwidth}
        \includegraphics[width=\linewidth]{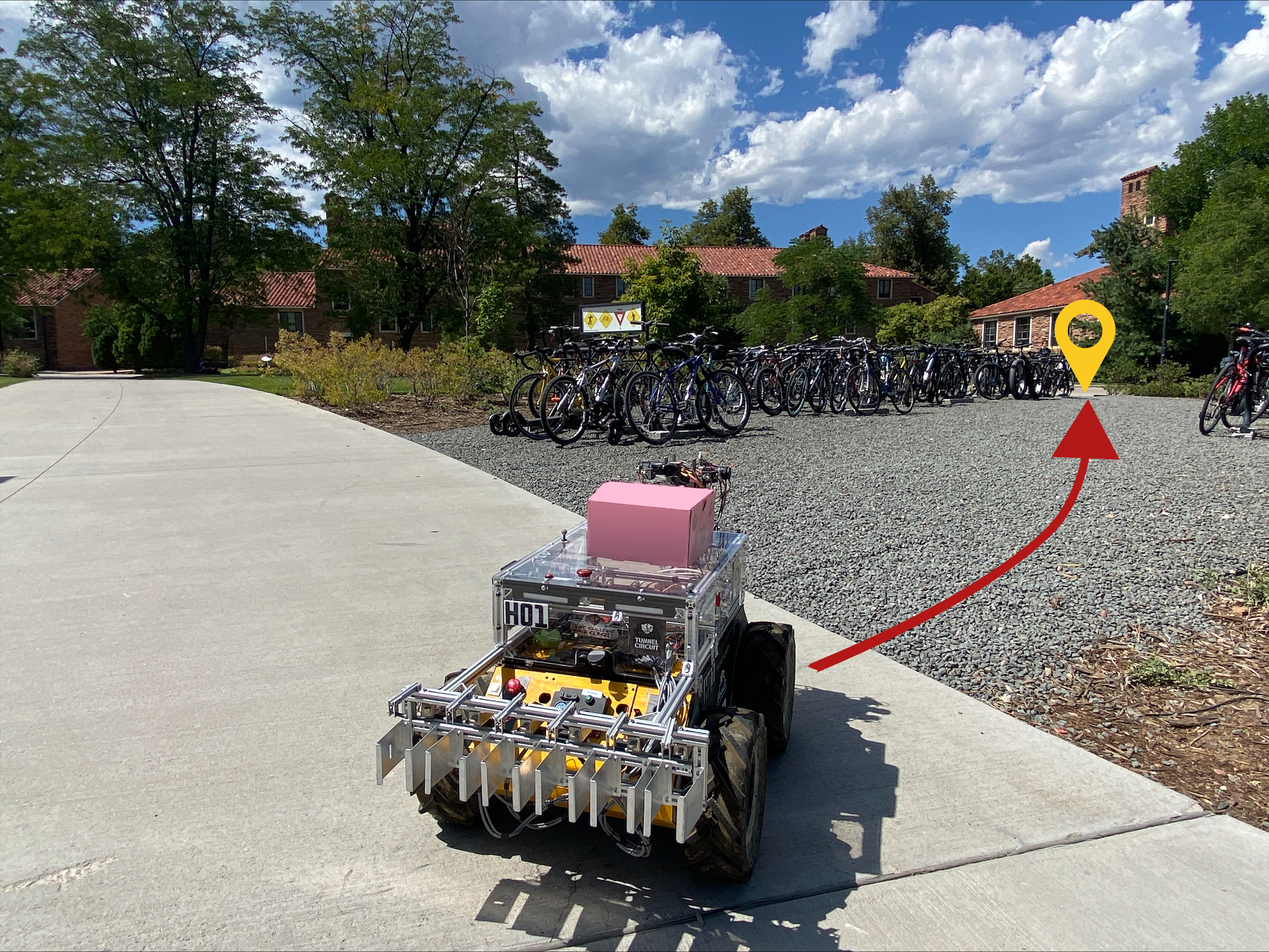} 
        \caption{User-aligned policy}
    \end{subfigure}
    \caption{The delivery robot above (goal: yellow pin) optimizes its path by considering factors like travel time and terrain types. Through active preference learning, it infers human rewards from trajectory rankings. (a) However, current state-of-the-art methods often struggle to efficiently generate informative trajectory pairs for these queries, leading to suboptimal results. To overcome this, CRED incorporates two key contributions: (b) counterfactual reasoning, which explores varied hypothetical preferences to produce more diverse trajectories, and (c) environment design, which ``imagines'' different scenarios--e.g. altering terrain from grass to gravel--to enhance the system's generalization capabilities. (d) As a result, the robot aligns to human preferences when deployed.}
    \label{fig:intro_figure_condensed}
\end{figure}

\section{INTRODUCTION}

Robots deployed in the real world must operate in diverse, unpredictable environments and interact with users whose preferences and expectations vary widely. Predefining appropriate behaviors for all possible scenarios is infeasible--robot behavior must instead be fine-tuned through human feedback to ensure alignment with user expectations. For example, in autonomous delivery, different robot embodiments (e.g., wheeled vs. legged robots) may need to learn which terrains are acceptable to traverse based on both physical capabilities and user-specified trade-offs between efficiency and risk \cite{mao2025pacer}. In domestic settings, users may prefer that household robots complete tasks more quickly, even if it means compromising slightly on task quality, such as folding laundry faster at the expense of producing perfectly neat folds. These types of trade-offs are often subtle, context dependent, and difficult to encode manually, underscoring the need for systems that can efficiently learn from user feedback during deployment.

Preference learning (PL) infers a reward function based on human-provided rankings of trajectories, eliminating the need to manually define rewards and allowing for personalization. Unlike inverse reinforcement learning, which depends on demonstration quality, PL allows non-expert users to provide input in complex domains such as Atari games and robot navigation \cite{christiano2017deep, zhang2025creste}. A key challenge, however, is that accurate reward learning often requires numerous preferences, limiting scalability to high-dimensional problems. To address this, active preference learning (APL) finds preference queries--sets of robot trajectories presented to a human--that maximize information gain \cite{Sadigh2017ActivePL, biyik2018batch}. Prior work makes this optimization tractable by restricting the query set to pre-generated trajectories \cite{erdem2020asking} or replay buffers \cite{lee2021pebble}, but this confines search to previously observed behaviors and may miss rare but informative scenarios needed to identify the true reward function. 

We introduce CRED, a novel and efficient query generation method for APL that learns reward functions which generalize to different scenarios by using 1) \textbf{C}ounterfactual \textbf{R}easoning and 2) \textbf{E}nvironment \textbf{D}esign. CRED's counterfactual reasoning generates queries that reflect different hypothesized human preferences instead of generating queries through random rollouts \cite{erdem2020asking}. Assuming a linear human reward model $R(\xi) = w^T\Phi(\xi)$ (the dot product of reward weights $w$ and trajectory features $\Phi(\xi)$), learning the reward function simplifies to learning $w$. We use Bayesian inference \cite{ramachandran2007bayesian} to maintain a belief over $w$, updated with each human query. By sampling diverse $w$ values (based on cosine similarity) from the current belief, CRED directly evaluates different human preferences and performs counterfactual reasoning, effectively asking ``what if $w_i$ or $w_j$ were the true reward weight?" Our second key insight is that the environment itself strongly influences the generated trajectories and thus can unnecessarily bound the information value of queries. CRED employs Bayesian Optimization \cite{bayesianopt2014} to efficiently find \emph{environment parameters} that yield the most informative preference queries.

The main contributions of this paper include a novel approach to preference learning, CRED, that leverages environment design and counterfactual reasoning. Empirical evaluations across close proximity (tabletop handover) and large length-scale (delivery navigation) tasks demonstrate that CRED's queries achieve higher reward accuracy and help users converge faster to the true reward function than prior methods. Moreover, user studies indicate that participants report lower mental workload and express stronger preference when interacting with CRED. 

\section{Related Works}
\textbf{Learning from Preferences.}
Preference learning infers a human's reward function by iteratively asking them to choose between robot trajectories \cite{erdem2020asking}. In active preference learning (APL), the objective is to identify the most informative query by selecting the one that maximizes the expected difference between the prior and the posterior belief distributions over reward functions \cite{Sadigh2017ActivePL}. Subsequent work explores different query selection objectives: maximize mutual information of the query and the estimated weights \cite{erdem2020asking}, maximize regret \cite{wilde2020active}, and maximize uncertainty in reward predictions \cite{lee2021pebble}. For example, Lee et al.~\cite{lee2021pebble} proposes ensemble-based sampling, which selects trajectory pairs with high disagreement across an ensemble of learned reward models, and entropy-based sampling, which maximizes the entropy of the preference distribution. In our paper, we focus on maximizing mutual information, as it directly quantifies the expected reduction in uncertainty over the reward weights and aligns well with Bayesian preference learning. However, our framework is general and can be extended to support other objectives.

A major challenge of APL is generating trajectory pairs that effectively optimize the chosen query selection objective. Prior methods such as \cite{erdem2020asking} and \cite{lee2021pebble} address this by selecting queries from a fixed dataset or replay buffer, but this restricts the diversity of queries and does not scale well to long-horizon tasks. Moreover, existing approaches typically generate queries within a single, fixed environment, limiting their ability to explore informative regions of the feature space. In contrast, our method leverages counterfactual reasoning to synthesize trajectories from a belief distribution over reward weights and jointly optimizes environment parameters. This combination enables the generation of more informative and diverse preference queries.

\textbf{Terrain Preferences for Navigation.}  
Prior work has incorporated human terrain preferences into robot navigation tasks by learning which surfaces to traverse or avoid. Karnan et al.~\cite{karnan2024wait} learn a visual representation space to extrapolate preferences to out-of-distribution terrains, Mao et al.~\cite{mao2025pacer} train a bird’s-eye view costmap from human feedback, and Zhang et al.~\cite{zhang2025creste} extend this by using demonstrations and counterfactuals to solicit additional human annotations. We adopt this domain for evaluation, but unlike these approaches which emphasize representation and costmap learning, our method targets the active selection of informative preference queries. As such, our framework can serve as a complementary front-end to generate the initial preference data that these methods require. 

\textbf{Environment Design in RL and Robotics.}
Environment design treats environment parameters as optimizable variables. In RL, it has been leveraged for curriculum learning to improve generalization and convergence, for instance, through co-evolution of agents and environment difficulty \cite{wang2019paired} or by modifying parameters to maximize an agent's learning potential \cite{dennis2020emergent}. In human-robot interaction, environment modification has been used to generate interpretable robot behaviors \cite{kulkarni2020designing}, legible human motion \cite{tung2024workspace, Tung2023HRI}, and to support collaborative teaming in settings like warehouse design \cite{ijcai2023p611} or tabletop reorganization \cite{bansal2020supportive}. Most relevant to our setting is the work of \cite{buening2024environment}, which formulates environment design for inverse reinforcement learning (IRL) as a bilevel optimization problem over transition dynamics to elicit more informative demonstrations from experts. Our work adapts this idea to preference learning, where feedback comes from comparisons rather than demonstrations. We apply environment design within APL to identify informative environments that improve query informativeness and reduce posterior uncertainty over the reward function.

\begin{figure*}[ht!]
    \centering
    \includegraphics[width=\linewidth]{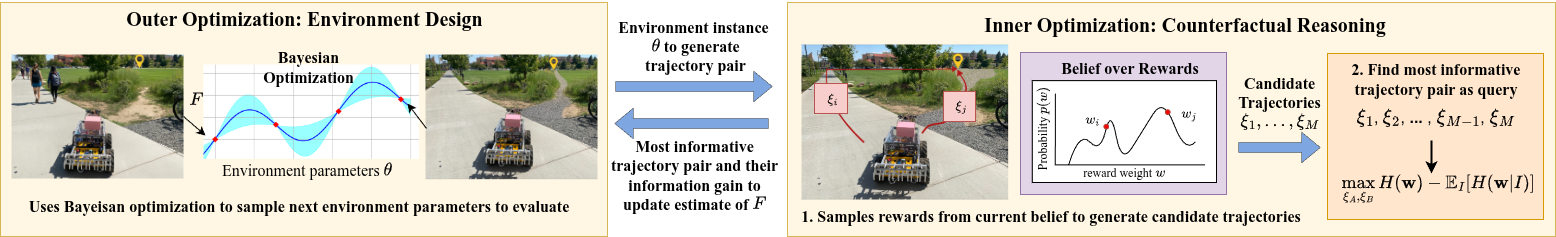}
    \caption{System overview of CRED as a bilevel optimization problem. \textbf{Outer optimization (Environment Design):} Bayesian optimization selects environment parameters $\theta$ to evaluate, seeking those that maximize estimates of query informativeness $F$. \textbf{Inner optimization (Counterfactual Reasoning):} Given $\theta$, the system samples reward weights from the current belief to generate candidate trajectories $\{\xi_1, \dots, \xi_M\}$. The most informative trajectory pair is returned to the outer optimization. }
    \label{fig:system-diagram}
\end{figure*}

\section{Preliminaries}
\textbf{Model.} We consider a fully observable environment modeled as a Markov decision process (MDP) consisting of $\{S, A, T, R, \gamma, S_0\}$, where $S$ is the set of states, $A$ is the set of actions, $T: S \times A \times S \rightarrow [0, 1]$ is the transition function, $R: S \times A \rightarrow \mathbb{R}$ is the reward function, $\gamma \in [0, 1)$ is the discount factor, and $S_0$ is the initial state distribution. Let $s_t \in S$ and $a_t \in A$ denote the state and action at time step $t$. A trajectory $\xi \in \Xi$ is a finite sequence of state-action pairs: $\xi = \left((s_0, a_0), (s_1, a_1), \dots, (s_H, a_H)\right)$ where $H$ is the planning horizon. We overload the notation $R$ and define the total reward of a trajectory as the sum of discounted per-step rewards: $R(\xi) = \sum_{t=0}^H \gamma^t R(s_t, a_t)$.

We assume a linear reward model over trajectory features: $R(\xi) = w^T \Phi(\xi)$
where $w \in \mathbb{R}^d$ are the unknown reward weights, and $\Phi(\xi) \in \mathbb{R}^d$ is the feature vector for trajectory $\xi$. To construct $\Phi(\xi)$, we define features over individual state-action pairs: let $\phi(s_t, a_t) \in \mathbb{R}^d$ be the feature vector at timestep $t$. The trajectory-level feature vector is then the discounted sum of per-step features: $\Phi(\xi) = \sum_{t=0}^H \gamma^t \phi(s_t, a_t)$.
Once the reward weights $w$ are learned from human feedback, we can optimize trajectories with respect to the learned reward function using reinforcement learning \cite{sutton2018reinforcement} or trajectory optimization methods such as CHOMP \cite{zucker2013chomp}.


\textbf{Preference learning.} The objective of preference learning is to learn $w$ by querying a human for their preferences between pairs of trajectories. A preference query typically asks ``Do you prefer trajectory $\xi_A$ or $\xi_B$?" \cite{biyik2018batch}. If a human prefers $\xi_A$ over $\xi_B$, it implies $R(\xi_A) > R(\xi_B)$, or equivalently $w^T\Phi(\xi_A) > w^T\Phi(\xi_B)$. From this strict inequality, we can derive that $w^T(\Phi(\xi_A)-\Phi(\xi_B)) > 0$. The human's preference $I$ can be encoded by $I = sign(w^T(\Phi(\xi_A)-\Phi(\xi_B)))$.

Human input can exhibit variability due to uncertainty in expressed preferences, which can be modeled using Boltzmann rationality. Under this model, the likelihood of a preference (Eqn.~\ref{eqn: likelihood}) is given by a softmax function over the reward values of the queried trajectories.
\begin{align}
    P(I \mid \mathbf{w}) &= \begin{cases} 
    \frac{\exp(R(\xi_A))}{\exp(R(\xi_A)) + \exp(R(\xi_B))} & \text{if } I = +1 \\
    \frac{\exp(R(\xi_B))}{\exp(R(\xi_A)) + \exp(R(\xi_B))} & \text{if } I = -1
    \end{cases}
    \label{eqn: likelihood}
\end{align}

Let $p(w)$ be our current belief distribution of the reward weights. We can perform a Bayesian update to compute the posterior given human input $I$, $p(w | I) \propto p(I | w)p(w)$. For uniqueness, we constrain the norm of the reward weights such that $\lVert w \rVert_2 \leq 1$. Since $p(w)$ can have arbitrary shapes, we use an adaptive Metropolis algorithm \cite{bj1080222083} to learn the posterior distribution. While we can use domain knowledge to initialize a non-uniform prior over reward weights, we adopt a uniform prior in all of our experiments for generality. Based on \cite{erdem2020asking}, the algorithm presents the human with a preference query and updates the belief distribution of $w$ until a fixed number of iterations is reached.

\textbf{Active Synthesis of Preference Queries.} To learn $w$ efficiently using minimal queries, active learning methods select preference queries $(\xi_A, \xi_B)$ that maximize information gain. This is equivalent to maximizing the mutual information between the query and the estimated weights $w$ \cite{erdem2020asking}. Our objective function $f$ is
\begin{equation}
    \max_{\xi_A, \xi_B} f(\xi_A, \xi_B) = \max_{\xi_A, \xi_B} H(\mathbf{w}) - \mathbb{E}_{I}[H(\mathbf{w} | I)]
\label{eqn: mutual information}
\end{equation}
where $H(w) = -\mathbb{E}_w[ \allowbreak log(p(w))]$ is the information entropy of the belief $p(w)$. This objective finds preference queries such that the difference between the entropy of the prior and the posterior is maximized. However, directly optimizing this objective is challenging in practice: the search space is non-convex, and optimization often converges to local maxima, resulting in poor sample efficiency. Section \ref{sec: technical} discusses our approach of using counterfactual reasoning and environment design to more effectively generate trajectories that optimize this information gain objective.

\textbf{Environment Design.} In traditional preference learning, the environment is fixed. However, the transition dynamics determine which trajectories are feasible and which features are observable. Consequently, the informativeness of a preference query depends not just on the trajectories, but also on the environment in which they are embedded.

We formalize the problem of environment design for preference learning as the selection of a sequence 
of environment parameters $\theta \in \Theta$, where each $\theta$ defines a specific transition function 
$T_\theta \in \mathbb{T}$. These parameters capture aspects of the environment such as terrain 
types, obstacle placements, or object configurations, and they determine which trajectories are feasible 
and what features can be observed.

For a given environment $\theta$, a trajectory $\xi$ is generated by executing a policy under the transition 
function $T_\theta$. Its features are represented by a function $\Phi(\xi, \theta)$, which depends on both the trajectory's state-action sequence and the environment. For example, 
the same path may traverse different terrain types depending on $\theta$, resulting in different features. 
This makes the feature map $\Phi(\cdot, \theta)$ environment-dependent and highlights the role of 
environment design in shaping the information content of preference queries.

At each round $i$, the learner (robot):
\begin{enumerate}
    \item Selects an environment $\theta^{(i)} \in \Theta$,
    \item Generates a pair of trajectories $(\xi_A^{(i)}, \xi_B^{(i)})$ in $T_{\theta^{(i)}}$,
    \item Queries the human: ``Which trajectory do you prefer?”,
    \item Updates its belief over the reward weights $\mathbf{w}$.
\end{enumerate}

This process allows the learner to actively shape the learning problem by selecting environments 
that yield the most informative trajectory comparisons for inferring the human's reward function.

\section{Technical Approach} \label{sec: technical}
Active preference learning faces challenges in generating trajectories that optimize for information gain (Eqn. \ref{eqn: mutual information}), as the objective function involves a pair of trajectories as variables. This task is further complicated by the fact that the optimization is typically constrained to a single environment, which may not adequately represent the full feature space, resulting in learned rewards that may fail to generalize effectively and suffer from sample inefficiency. Our approach using counterfactual reasoning and environment design to address these issues is summarized in Fig. \ref{fig:system-diagram}.

\subsection{Counterfactual Reasoning} \label{sec: counterfactual}
Counterfactual reasoning explores different trajectories that could result if a hypothesized set of reward weights were the true weights. We maintain a belief over the weights (via an adaptive Metropolis algorithm \cite{bj1080222083}) while estimating the human's reward function, where each sample of weights could lead to a different policy and consequently different trajectories when the policy is executed. This allows us to pose counterfactual questions, such as ``what if reward $i$ is the true reward as opposed to reward $j$?" Let $w_i$ be an instance of reward weights sampled from our belief. We can use reinforcement learning (RL) \cite{sutton2018reinforcement} to train a policy $\pi_i$ that maximizes the reward function induced by $w_i$. Rolling out $\pi_i$ in a given environment yields a trajectory $\xi_i$. Alternatively, we can directly optimize the trajectory with respect to $w_i^T \Phi(\xi)$ using trajectory optimization methods such as CHOMP \cite{zucker2013chomp} (\cref{alg: counterfactual} lines \ref{cr: generate traj}).

By sampling reward weights and generating trajectories, we construct a set of counterfactual trajectories that represent different human preferences. We then evaluate the information gain objective $f$ (\cref{eqn: mutual information}) for each pair of trajectories to identify the most informative preference query (\cref{alg: counterfactual} lines \ref{cr: infogain}-\ref{cr: max}). To minimize the number of evaluations of the objective function, we start by sampling $N$ reward weights. We then select the most diverse $M$ weights, where $M < N$, from this set by sequentially computing diversity based on cosine similarity, forming our final set of reward weights for evaluation (\cref{alg: counterfactual} lines \ref{cr: sample}-\ref{cr: diverse weights}).
\begin{algorithm}[t]
\caption{Counterfactual Reasoning}
\begin{algorithmic}[1]
\Require Belief $P(w)$, $N$ samples, $M$ subset size

\State Sample $\{w_1, \dots, w_N\} \sim P(w)$ \label{cr: sample}
\State Select $M$ diverse weights (e.g., max cosine distance) \label{cr: diverse weights}
\For{each selected $w_k$}
    \State Generate trajectory $\xi_k$ by maximizing $w_k^\top \Phi(\xi)$
    \par\hspace*{0pt}\text{// e.g., via RL or CHOMP} \label{cr: generate traj}
\EndFor
\State Compute information gain $f$ (Eq. \ref{eqn: mutual information}) for all pairs $\xi_i, \xi_j$  \label{cr: infogain}
\State Return most informative pair $(\xi_i, \xi_j)$ \label{cr: max}
\end{algorithmic}
\label{alg: counterfactual}
\end{algorithm}
\begin{algorithm}
\caption{Environment Design}
\begin{algorithmic}[1]
\Require Environment parameters $\Theta$, Bayesian optimization iterations $N$
\For{$t = 1$ to $N$} 
    \State Propose $\theta^t$ using Bayesian optimization
    \State Generate $(\xi_A^{(t)}, \xi_B^{(t)})$ via CR (\Cref{alg: counterfactual}) in env $\theta^t$
    \State Compute information gain $F(\xi_A, \xi_B; \theta^t)$
    \State Update GP model with $(\theta^t, F(\xi_A, \xi_B; \theta^t))$
\EndFor
\State Return optimal $\theta^*$ found and corresponding $(\xi_A, \xi_B)$
\end{algorithmic}
\label{alg: environment design}
\end{algorithm}
\subsection{Environment Design} \label{sec: env design}

While counterfactual reasoning generates trajectory pairs optimizing for different reward weights, the fixed environment can limit their ability to reveal crucial preference distinctions. We posit that if we have the ability to ``imagine'' new environments, we can better generate trajectories that show the differences between the different reward weights.

To formalize this idea, we make explicit the role of environment parameters in trajectory generation. Recall that $\Theta$ denotes the set of configurable environment parameters where $\theta \in \Theta$ induces a transition function $T_\theta$. The feature function $\Phi$ is environment-dependent and written as $\Phi(\xi, \theta)$. Let $F(\xi_A, \xi_B; \theta)$ denote the information gain from presenting the trajectory pair $(\xi_A, \xi_B)$ in environment $\theta$, defined analogously to Eqn.~\ref{eqn: mutual information}, but accounting for the dependence of trajectories and features on $\theta$. We formulate environment design as a bilevel optimization problem \footnote{
In principle, environment parameters $\theta$ and trajectory pairs $(\xi_A, \xi_B)$ 
could be jointly optimized in a single-level formulation. However, the feasible set of 
trajectories $\Xi(\theta)$ depends on $\theta$ through the induced transition function 
$T_{\theta}$, making joint optimization computationally expensive. The bilevel formulation leverages this structure by first selecting $\theta$ and 
then optimizing over $\Xi(\theta)$, which is more tractable in practice.
}:
\begin{equation}
    \max_{\theta \in \Theta} \max_{(\xi_A, \xi_B) \in \Xi(\theta)} F(\xi_A, \xi_B; \theta) \label{eqn: bilevel opt}
\end{equation}
where $\Xi(\theta)$ denotes the set of feasible trajectory pairs under the MDP with transition 
$T_{\theta}$. The outer optimization selects the environment parameters $\theta$ that maximize the informativeness of the resulting preference query, while the inner optimization identifies the most informative trajectory pair $(\xi_A, \xi_B)$ that can be generated in that environment. This bilevel structure reflects the coupling between environment design and query synthesis in our framework.



Since $F(\xi_A, \xi_B; \theta)$ is generally not differentiable with respect to $\theta$, we use Bayesian optimization \cite{snoek2012practical}, which models $F$ with a Gaussian process (GP) defined by a mean function $m: \Theta \rightarrow \mathbb{R}$ and a positive definite covariance function $K: \Theta \times \Theta \rightarrow \mathbb{R}$. We use the upper confidence bound (UCB) acquisition function, selecting $\theta$ that maximizes $UCB(\theta) = \mu(\theta) + \kappa \sigma(\theta)$, where $\kappa$ balances exploitation against exploration. We use the \texttt{BayesianOptimization} Python library \cite{bayesianopt2014} which employs a Matérn kernel with hyperparameters fit via maximum likelihood estimation. Unless otherwise specified, we use the library’s default settings. Algorithm \ref{alg: environment design} shows the pseudocode for environment design.
\begin{figure*}[!ht]
\centering
\resizebox{0.92\textwidth}{!}{%
\begin{minipage}{\textwidth}
\centering
\begin{subfigure}{0.3\textwidth}
  \centering
  \includegraphics[width=\linewidth]{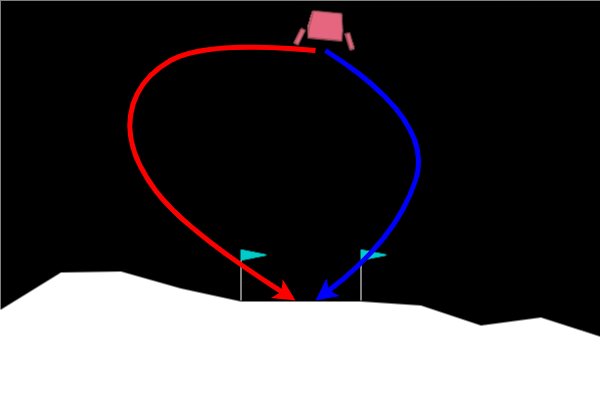}
\end{subfigure}
\hfill
\begin{subfigure}{0.3\textwidth}
  \centering
  \includegraphics[width=\linewidth]{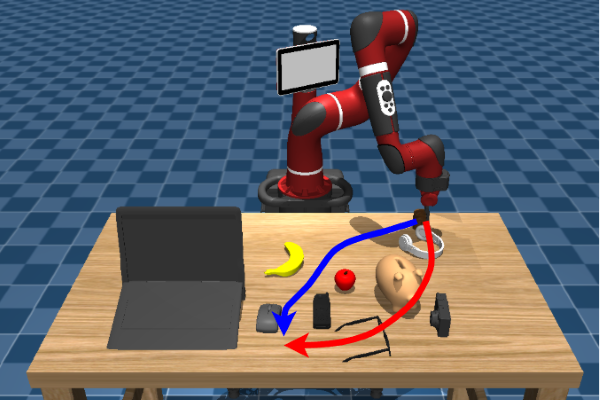}
\end{subfigure}
\hfill
\begin{subfigure}{0.3\textwidth}
  \centering
  \includegraphics[width=\linewidth]{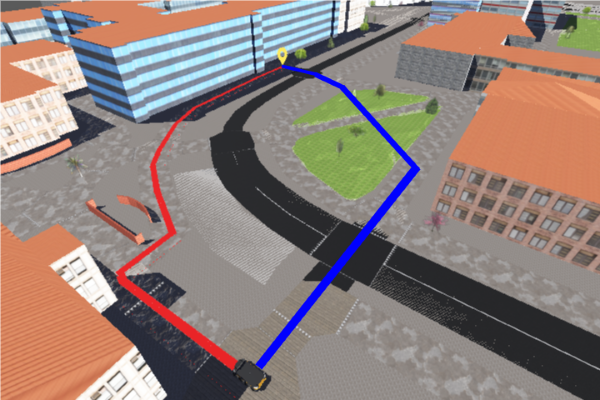}
\end{subfigure}

\vspace{1em}

\footnotesize
\begin{tabular}{@{}>{\centering\arraybackslash}p{0.1\textwidth} 
    >{\centering\arraybackslash}p{0.25\textwidth} 
    >{\centering\arraybackslash}p{0.3\textwidth} 
    >{\centering\arraybackslash}p{0.3\textwidth}@{}}
\toprule
 & \textbf{Lunar Lander} & \textbf{Table Top} & \textbf{Navigation} \\
\midrule
\textbf{Features $\Phi$} &
vertical speed, horizontal position &
objects that the end effector hovers over and the trajectory length &
path length, terrain traversed (paved, grass, asphalt and concrete)\\
\midrule
\textbf{Env params $\theta$} &
wind power, ranging from 1 to 20 &
object positions on the table &
terrain type of road network \\
\midrule
\textbf{Preferences} &
approach direction, landing speed &
objects to avoid hovering over &
terrains to avoid and path length \\
\bottomrule
\end{tabular}
\end{minipage}
} 
\caption{Examples of preference queries with details and visualization across the three simulation domains. 
}
\label{fig:pref_queries}
\end{figure*}

\section{Experiments}

We evaluate CRED across a suite of simulation experiments and a user study. Through these experiments we demonstrate that CRED enables more accurate and efficient inference of human reward functions than existing methods and characterize the contributions of its core components.

We begin with simulation experiments in three domains (Lunar Lander, Tabletop Manipulation, and Navigation) where we measure the accuracy of learned rewards and compare CRED against two established baselines \cite{erdem2020asking, christiano2017deep}. 
We conduct ablation studies to isolate the impact of counterfactual reasoning and environment design, followed by a comparison of our environment design strategy with domain randomization, an alternative that samples environments uniformly at random. 
Finally, we present user study results evaluating if real users can effectively teach preferences using CRED in manipulation and navigation tasks. 

Our experiments address the following hypotheses:
\begin{itemize}
    \item \textbf{H1:} CRED achieves a higher reward accuracy with fewer preference queries compared to baselines \cite{erdem2020asking, christiano2017deep}. 
    \item \textbf{H2:} Both counterfactual reasoning and environment design contribute significantly to CRED’s performance
    \item \textbf{H3:} CRED outperforms domain randomization by generating more informative environments that accelerate reward learning. 
    \item \textbf{H4:} Real users prefer interacting with CRED over baseline methods in practical manipulation and navigation tasks and report lower mental workload as measured by NASA-TLX.
\end{itemize}

\subsection{Simulation Experiments}

\begin{figure*}[!t]
\centering
\renewcommand{\arraystretch}{1.2}
\setlength{\tabcolsep}{5pt}

\begin{tabular}{c c c c}
\hline
 & \textbf{Lunar Lander} & \textbf{Tabletop} & \textbf{Navigation} \\
\hline
\rotatebox{90}{\hspace{16pt} \textbf{Baselines}} &
\includegraphics[width=0.3\textwidth]{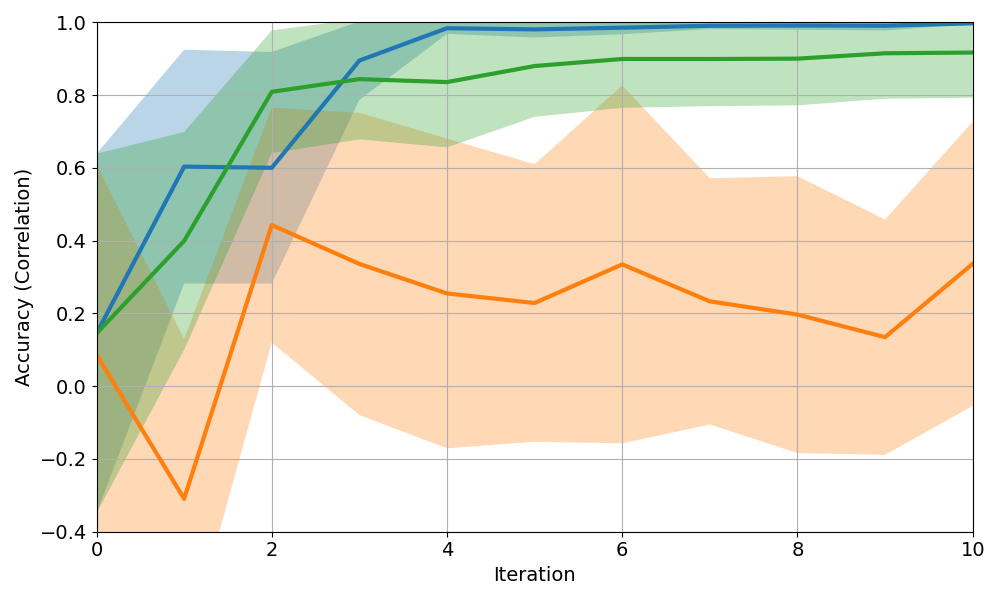} &
\includegraphics[width=0.3\textwidth]{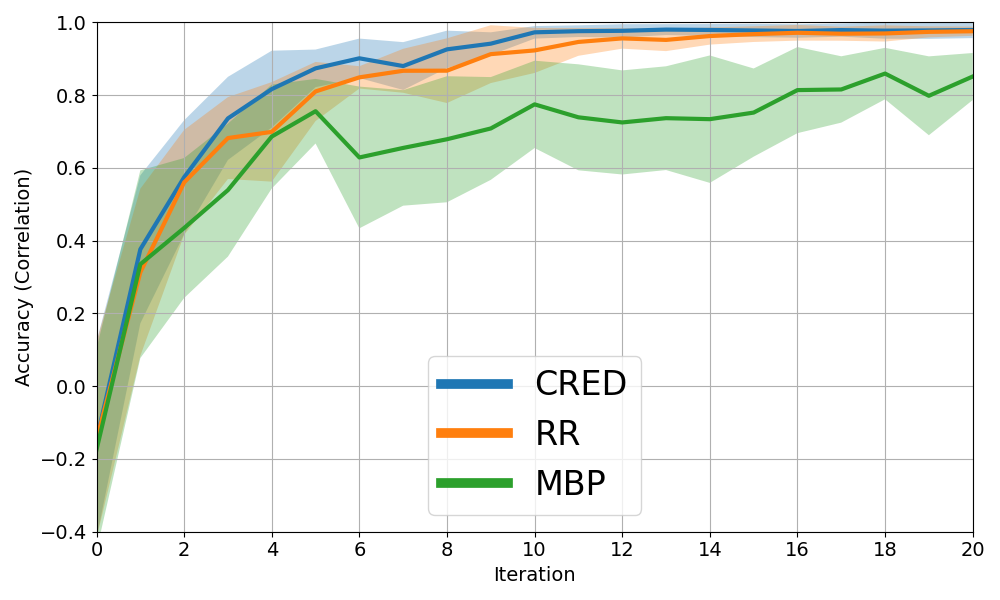} &
\includegraphics[width=0.3\textwidth]{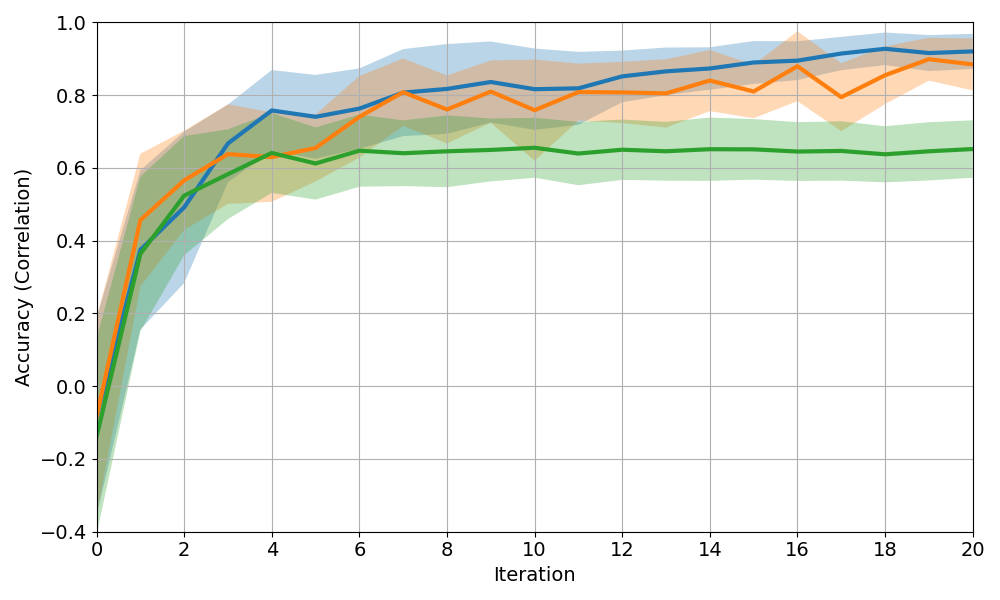} \\

\rotatebox{90}{\hspace{16pt} \textbf{Ablation}} &
\includegraphics[width=0.3\textwidth]{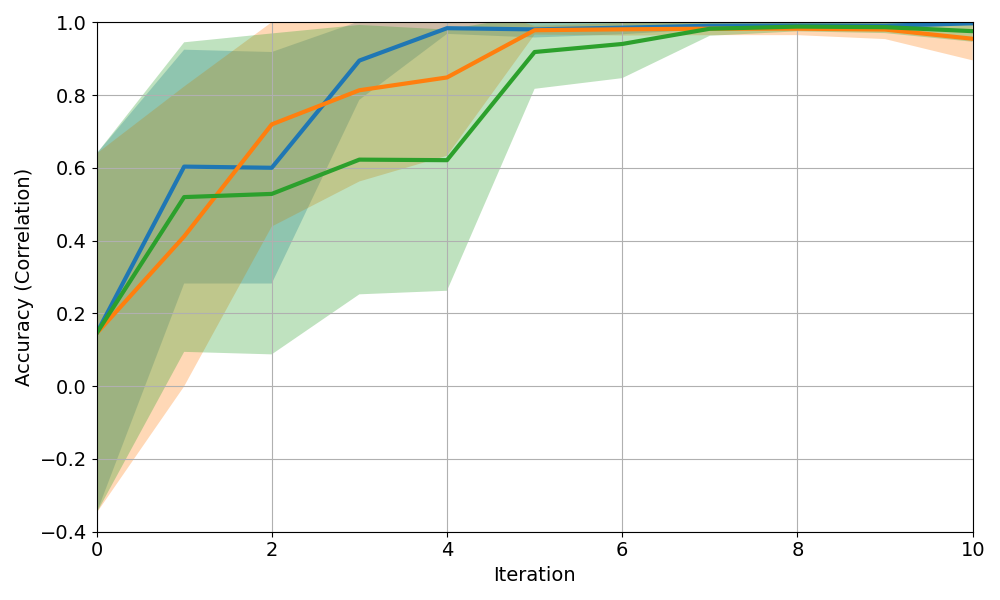} &
\includegraphics[width=0.3\textwidth]{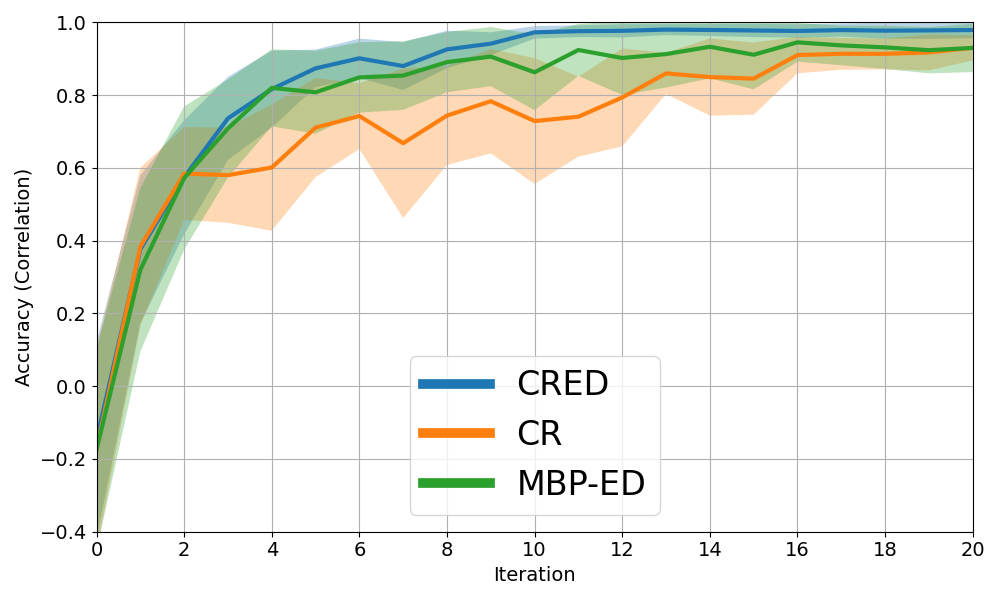} &
\includegraphics[width=0.3\textwidth]{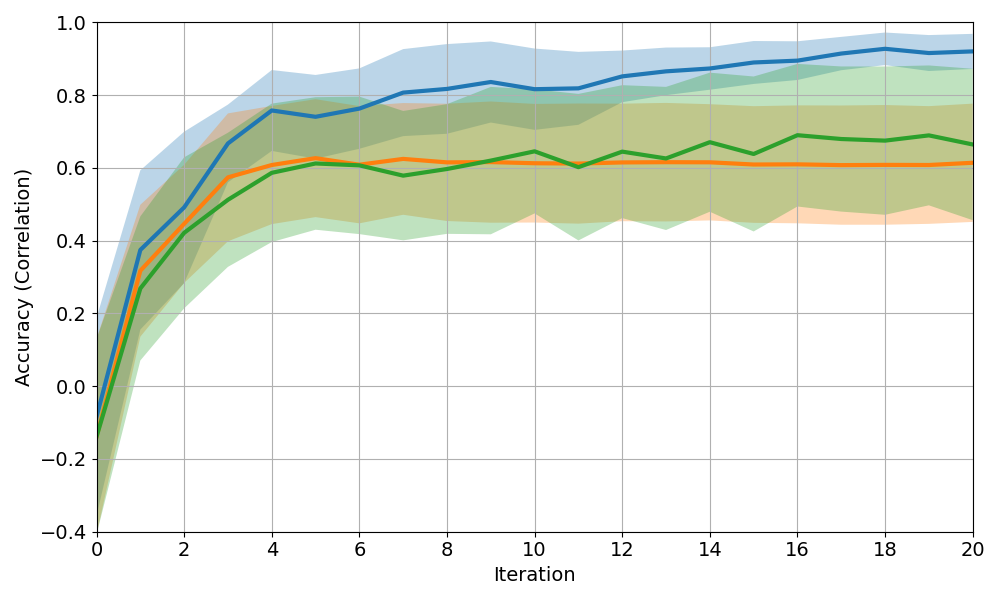} \\
\rotatebox{90}{\quad \textbf{Domain Rand.}} &
\includegraphics[width=0.3\textwidth]{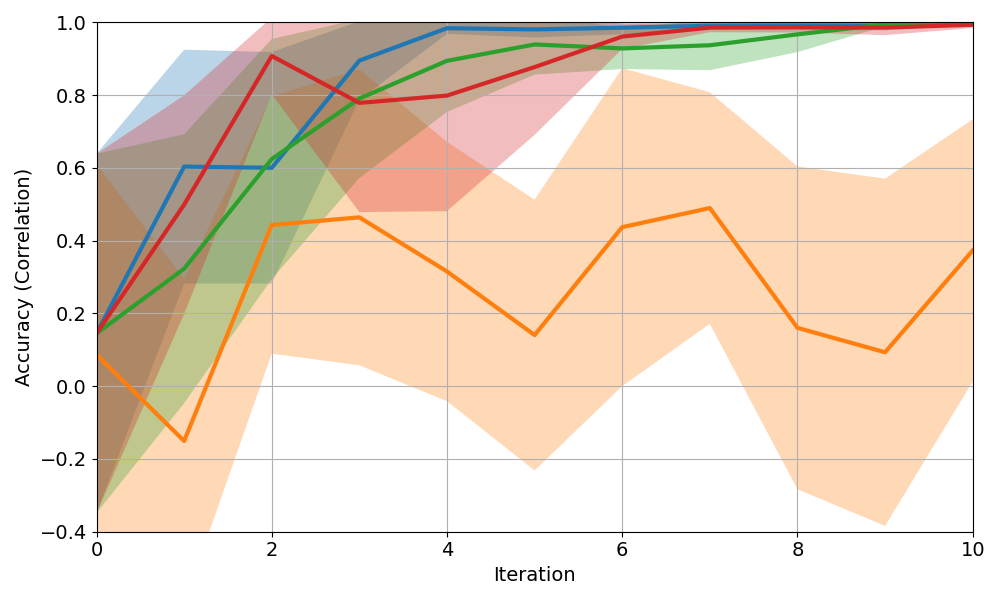} &
\includegraphics[width=0.3\textwidth]{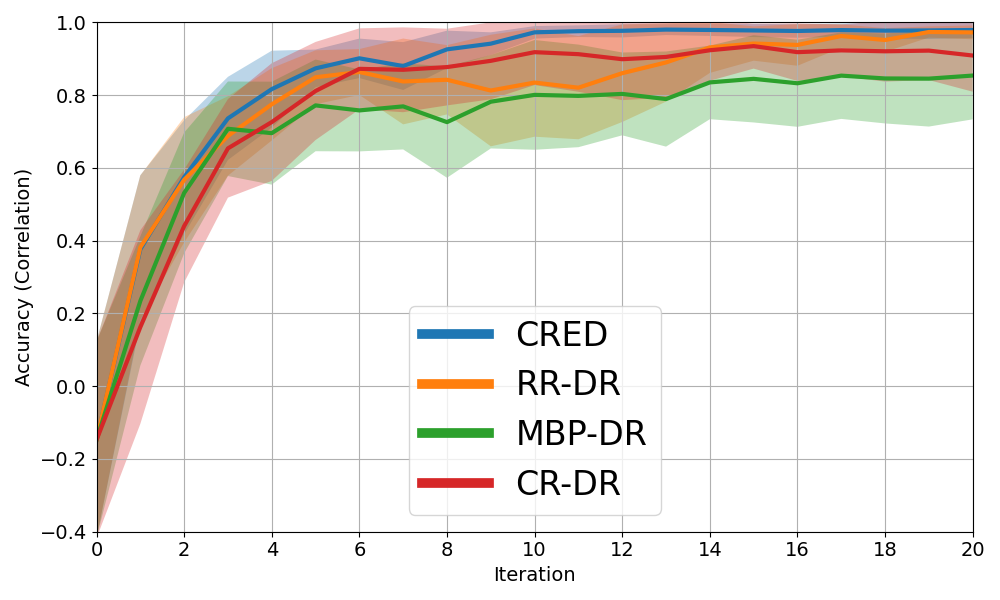} &
\includegraphics[width=0.3\textwidth]{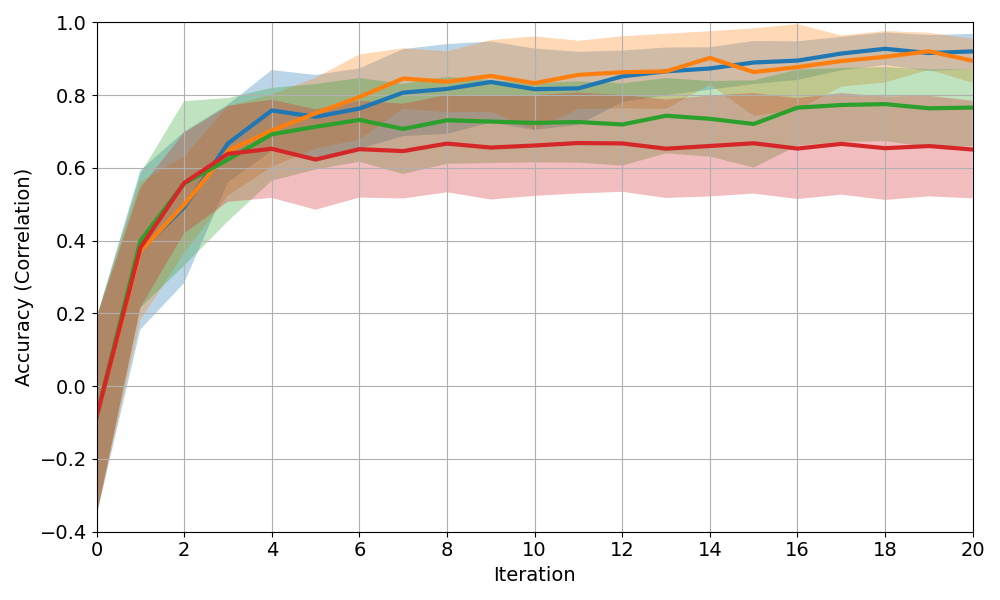} \\
\end{tabular}

\caption{Accuracy of the estimated rewards across three domains (columns) for each experiment (rows). 
}
\label{fig:results_table}
\end{figure*}

For all simulation experiments, we simulate $10$ users, each initialized with a distinct ground-truth reward weight vector. To encourage diversity, we sample $1000$ random weight vectors and select the cluster centers obtained via $K$-means as the ground-truth weights. This setup allows us to test whether CRED can generalize across heterogeneous user preferences. 

\subsubsection{Environment Setup}
\textbf{Lunar Lander}. We use the Gymnasium implementation of Lunar Lander \cite{towers2024gymnasium}, where a lander must safely reach a designated pad in a 2D environment. 
Beyond the environment's default objective of safe landing, we introduce user-dependent preferences such as approaching from the left versus right side and controlling descent speed. The feature vector $\Phi$ encodes the lander’s vertical velocity and horizontal position, while the environment parameter $\theta$ specifies the wind power applied during descent. To generate trajectories, we train PPO \cite{schulman2017proximal} agents using the Stable Baselines implementation. We use the default hyperparameters, with two exceptions: the target KL divergence is set to 0.01, and both the policy and value networks are two-layer MLPs with 256 hidden units per layer.

\textbf{Tabletop Manipulation}. We adapt a tabletop manipulation domain \cite{peng2025towards} in MuJoCo \cite{Todorov2012MuJoCoAP}, where a robot delivers a cup of coffee across a cluttered table. To avoid spills on critical items such as electronics, the robot must learn user preferences for safe trajectories. Objects on the table include fruits, laptop, a phone, glasses, headphones, a camera, and a piggy bank. The features encode how often the robot’s trajectory hovers above each object, while the environment parameters $\theta$ specify object placements. To make the search over possible object configurations tractable, we use a variational autoencoder (VAE) \cite{cinelli2021variational} to compress environments into a latent space $Z$ (see Appendix \ref{sec: tabletop vae}), where the environment parameters $\theta$ correspond to latent vectors $z \in Z$. We use CHOMP \cite{zucker2013chomp} to generate trajectories and augment the cost function with $-R$ to incorporate the rewards.

\textbf{Navigation}. We consider a delivery task where a robot must transport food to a customer by navigating between predefined start and goal locations over a street network derived from OpenStreetMaps \cite{OpenStreetMap}. The network is simulated in Webots \cite{Webots} for experiments and user studies. The features encode the path length and the proportion of each terrain type (e.g., asphalt, paved, grass, concrete, and brick) along the route. The environment parameters $\theta$ control the surface types assigned to the edges on the shortest path, thereby altering the trade-offs between efficiency and terrain preferences. Trajectories are generated using value iteration \cite{sutton2018reinforcement}, where graph nodes correspond to states and outgoing edges define the available actions at each node.

\subsubsection{Metrics}

We measure accuracy using the sample correlation coefficient $r$ between ground truth and estimated rewards:
\begin{equation}
r = \frac{\sum_{i=1}^n \left(R_{gt}^i - \bar{R}_{gt}\right)\left(R_{est}^i - \bar{R}_{est}\right)}
{\sqrt{\sum_{i=1}^n \left(R_{gt}^i - \bar{R}_{gt}\right)^2} \; \sqrt{\sum_{i=1}^n \left(R_{est}^i - \bar{R}_{est}\right)^2}},
\label{eq: correlation}
\end{equation}
where $R_{gt}^i$ and $R_{est}^i$ are the ground truth and estimated rewards for sample $i$, and $\bar{R}$ denotes the sample mean. We compute $R_{gt}$ and $R_{est}$ by evaluating the true and learned weight vectors over a grid of feature vectors, with feature ranges determined from a set of trajectories sampled in each environment. 

\subsubsection{Baselines}
We compare CRED against two state-of-the-art preference learning baselines. The first optimizes the mutual information objective (Eqn.~\ref{eqn: mutual information}) over pre-generated trajectories from random rollouts \cite{erdem2020asking}, which we denote as \textbf{RR}. The second, inspired by \cite{christiano2017deep} and used in later work such as PEBBLE \cite{lee2021pebble}, learns a reward function by training a neural network on preference data. In our Bayesian belief framework, we replicate this baseline by using the mean belief as the reward function. During rollouts, the policy trained on the learned reward function follows its action with probability $1-\epsilon$ and explores with probability $\epsilon$ ($\epsilon=0.25$ in our experiments). For CHOMP, exploration is simulated by adding Gaussian noise. We refer to this baseline as the \textbf{M}ean \textbf{B}elief \textbf{P}olicy (\textbf{MBP}). 

\begin{figure*}[htbp]
    \centering
    \begin{subfigure}[t]{0.24\textwidth}
        \centering
        \includegraphics[width=\linewidth]{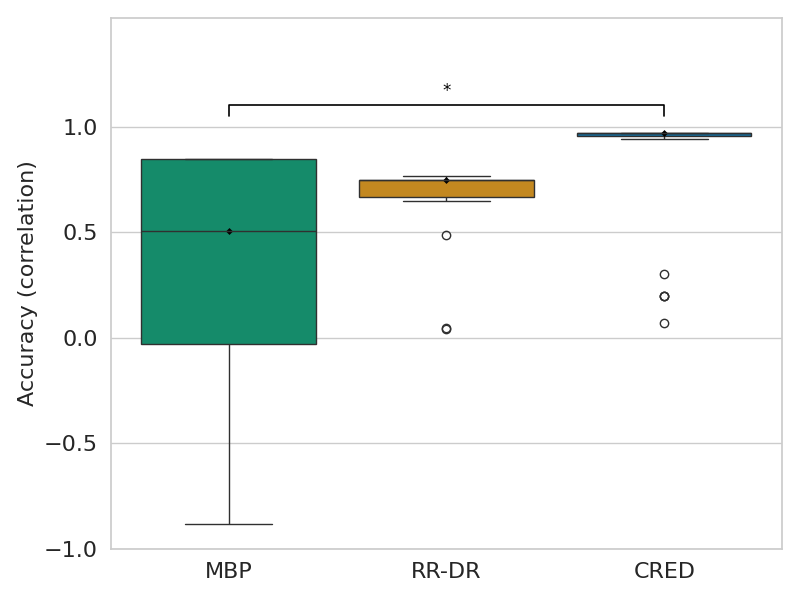}
        \caption{Tabletop accuracy}
        \label{fig:user-study-tabletop}
    \end{subfigure}
    \hfill
    \begin{subfigure}[t]{0.24\textwidth}
        \centering
        \includegraphics[width=\linewidth]{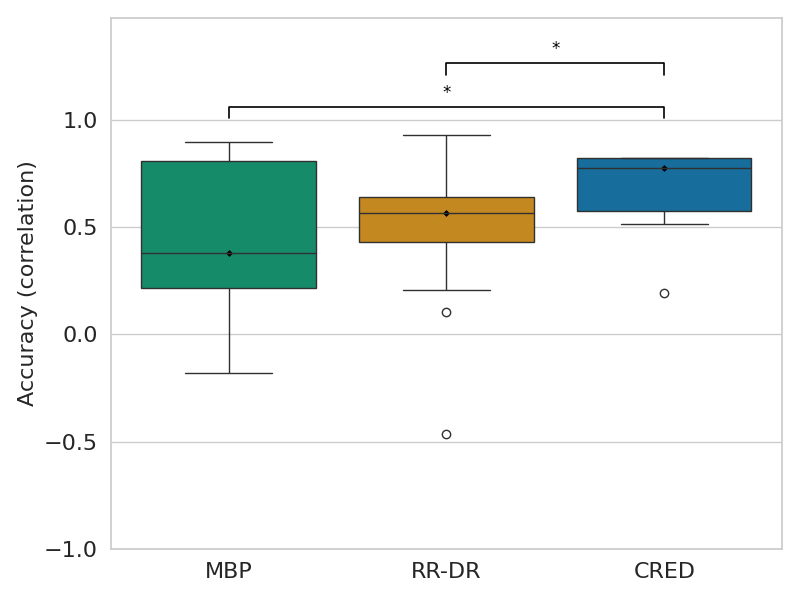}
        \caption{Navigation accuracy}
        \label{fig:user-study-navigation}
    \end{subfigure}
    \hfill
    \begin{subfigure}[t]{0.24\textwidth}
        \centering
        \includegraphics[width=\linewidth]{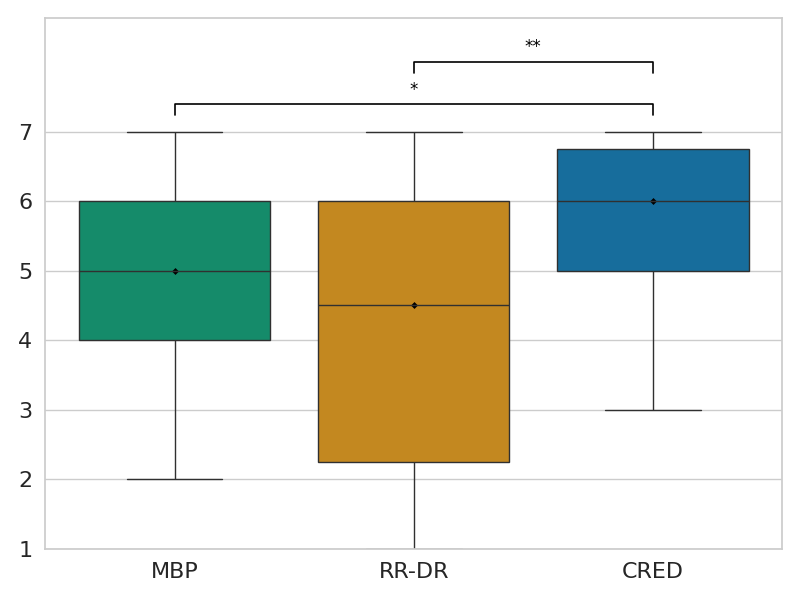}
        \caption{Ease of choice}
        \label{fig:user-study-ease-of-choice}
    \end{subfigure}
    \hfill
    \begin{subfigure}[t]{0.24\textwidth}
        \centering
        \includegraphics[width=\linewidth]{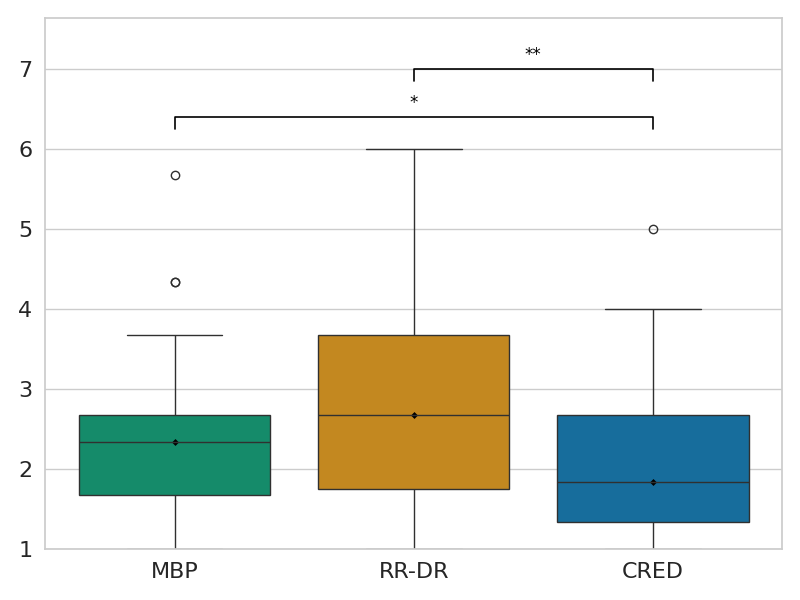}
        \caption{NASA-TLX}
        \label{fig:user-study-nasa-tlx}
    \end{subfigure}
    \caption{
    Results from the user study: higher is better for (a)-(c), lower is better for (d). Within each plot the central line denotes the median, the upper and lower edges correspond to the first (Q1) and third (Q3) quartiles, and the whiskers extend to 1.5 times the interquartile range from Q1 and Q3. Statistically significant differences indicated as: * $p \leq 0.05$, ** $p \leq 0.01$.}
    \label{fig:user-study-results}
\end{figure*}

\subsubsection{Simulation Results}

\textbf{Baseline Comparison}. Figure~\ref{fig:results_table} (row 1) reports mean accuracy with 95\% confidence intervals across all three domains, comparing CRED to RR \cite{erdem2020asking} and MBP \cite{christiano2017deep}. CRED consistently achieves the highest final accuracy, reaching $0.998$, $0.979$, and $0.906$ in Lunar Lander, Tabletop, and Navigation, respectively. CRED also converges rapidly, surpassing $95\%$ accuracy within $4$ iterations for Lunar Lander and $10$ for Tabletop. Navigation requires around $16$ iterations, reflecting the greater difficulty of distinguishing preferences across multiple terrain types. By contrast, MBP plateaus around $60\%$ accuracy and fails to produce informative queries after roughly $8$ iterations. Because MBP cannot get feedback in novel environments, it is restricted to the features available in the current environment, limiting its effectiveness. Overall, CRED achieves both higher reward accuracy and faster convergence, supporting \textbf{H1}.

\textbf{Ablations}. To assess the contribution of each component in CRED, we evaluate two ablations. 
First, \textbf{CR} removes environment design, relying only on counterfactual reasoning. 
Second, \textbf{MBP-ED} removes counterfactual reasoning, applying environment design to the mean-belief policy baseline. As shown in row 2 of Figure~\ref{fig:results_table}, both ablations yield either lower final accuracy or slower convergence across all domains, indicating that environment design and counterfactual reasoning are complementary. These results support \textbf{H2}.


We further test the environment design strategy against \textbf{domain randomization (DR)}, which selects environment parameters $\theta \in \Theta$ uniformly at random. For fairness, the number of environments sampled under DR matches the number evaluated during Bayesian optimization. We construct three DR baselines: \textbf{RR-DR}, which generates random rollouts in randomized environments; \textbf{MBP-DR}, which executes trajectories from the mean-belief policy under random environments; and \textbf{CR-DR}, which applies counterfactual reasoning without guided environment selection. Results (row 3 of Figure~\ref{fig:results_table}) show that all DR variants underperform CRED (with the exception of RR-DR in Navigation), highlighting the benefit of using Bayesian optimization to actively search for informative environments rather than sampling them uniformly at random (\textbf{H3}). We note that RR-DR does not enforce a goal-reaching constraint on trajectories, enabling greater diversity in the feature space. However, our user study (Sec.~\ref{sec:user_study}) shows that RR-DR trajectories are harder for users to interpret, limiting their practicality despite their competitive accuracy in simulation.

\subsection{User Study} \label{sec:user_study}

\subsubsection{Setup}
We conducted an IRB-approved user study to evaluate CRED in both tabletop and navigation tasks. A total of 25 participants were recruited from the university campus (ages 20--39, $M=26.8$, $SD=4.5$; 18 male, 6 female, 1 other). Participants were randomly assigned to conditions in a within-subjects design, with each participant completing 6 preference queries per task. We compared CRED against two baselines: MBP \cite{christiano2017deep} and RR-DR, which uses pre-generated trajectories \cite{erdem2020asking} augmented with domain randomization. We did not include MBP with domain randomization (MBP-DR), as simulation results showed that it performed comparably to MBP alone. To avoid participant fatigue, we restricted the study to these three conditions. The order of conditions was randomized for each task to mitigate ordering effects.

In each trial, participants were presented with a pair of robot trajectories and asked to select the one they preferred. Trajectories were generated in simulation and visualized as arrows overlaid on the task environment, similar to Fig. \ref{fig:pref_queries}. The reward functions inferred from these preferences were subsequently deployed on the physical robot platforms, as demonstrated in the supplementary video. For the tabletop task, the preferred behavior was to avoid passing over electronics, while for the navigation task, the preferred behavior was to avoid crossing asphalt roads. Participants were informed that although the robot could traverse grass, such routes may be less energy efficient. These task instructions were chosen to define a ground-truth reward function \footnote{Ground-truth rewards are defined as linear functions of trajectory features, normalized to unit-length weight vectors. For the tabletop task, 
features are the number of waypoints over \texttt{fruit}, \texttt{accessories}, and \texttt{electronics}, and the trajectory length, with weights $[-0.1, -0.1, -2.0, -1.0]$. For the navigation task, features are path length and the distances traversed on \texttt{paved}, \texttt{grass}, \texttt{asphalt}, and \texttt{concrete}, with weights $[-1.0, -0.1, -2.0, -5.0, -0.1]$.}, allowing us to evaluate the accuracy of the learned rewards while also capturing realistic trade-offs in user preferences.

After each condition, participants completed a short questionnaire. Subjective workload was measured using the NASA-TLX \cite{hart1988development} survey, and perceived ease of comparison was measured with a 7-point Likert item (“It was easy to choose between the trajectories that the robot showed me,” 1=strongly disagree, 7=strongly agree). To assess learning performance, we compared the inferred reward functions to the ground truth rewards by computing the reward correlation (Eq. \ref{eq: correlation}), which was also used in the simulation experiments.

\subsubsection{Results}

Fig. \ref{fig:user-study-results} shows the box plots of the user study results. For the tabletop task, CRED achieved higher reward correlation with the ground truth rewards (median $r = 0.97$, IQR $= 0.01$), outperforming RR-DR (median $r = 0.75$, IQR $= 0.08$) and MBP (median $r=0.51$, IQR $= 0.88$). Similarly, in the navigation task, CRED again showed better performance (median $r = 0.78$, IQR $= 0.25$), compared to RR-DR (median $r = 0.57$, IQR $= 0.21$) and MBP (median $r = 0.38$, IQR $= 0.59$). In terms of subjective results, participants reported lower mental workload under CRED (NASA-TLX: median $= 1.83$, IQR $= 1.33$) compared to DR-RR ($2.67, 1.92$) and MBP ($2.33, 1.00$). They also rated CRED higher on ease of comparison ($6.0, 1.75$) than DR-RR ($4.5, 3.75$) and MBP ($5.0, 2.00$), supporting \textbf{H4}.

Statistical comparisons were conducted using the Wilcoxon signed-rank test, a nonparametric test appropriate for ordinal measures (e.g., Likert ratings, NASA-TLX scores) and bounded non-normally distributed measures such as reward correlations. To account for multiple pairwise comparisons, we applied Holm–Bonferroni correction to control the family-wise error rate. CRED significantly outperformed both RR-DR and MBP in the tabletop and navigation tasks, with the exception of the CRED vs. RR-DR comparison in tabletop reward accuracy. 
We observed a few outliers in the tabletop task for both CRED and RR-DR, which may be attributable to participants not following task instructions. These results show that CRED more consistently recovers the true rewards across both navigation and manipulation domains. Furthermore, participants found CRED less mentally demanding and perceived the queries generated by CRED as easier to evaluate.

\section{Conclusion}

We introduce CRED, a query generation framework that jointly optimizes environments and trajectories for active preference learning. Experiments show that CRED achieves higher sample efficiency and reward accuracy than prior methods, and ablations highlight the importance of counterfactual reasoning, environment design, and Bayesian optimization for selecting informative environments. A user study in tabletop and navigation tasks demonstrates that CRED produces queries that are easier to answer and impose lower mental workload. CRED is applicable across policy classes (e.g., PPO, value iteration, trajectory optimization) and domains with different preferences. One limitation is the computational complexity: as a bilevel formulation, CRED requires an outer search over environment parameters and an inner search over trajectories. Future work can mitigate this cost through parallelization of environment evaluations and more efficient approximations of query informativeness.

\addtolength{\textheight}{-12cm}   



\section*{APPENDIX}

\subsection{VAE for Tabletop Object Positions} \label{sec: tabletop vae}

We learn a low-dimensional latent space $Z$ over tabletop object layouts with a convolutional VAE.
Each environment is represented as a $C{\times}H{\times}W$ binary grid $x$ (here $C{=}7$ object channels; $H{=}W{=}4$), where channel $c$ encodes the occupancy of object $c$ over the grid.
The encoder $q_\phi(z\!\mid\!x)$ consists of two stride-2 convolutions (32 and 64 filters, kernel $3{\times}3$, ReLU), followed by flattening and linear heads to the mean $\mu(x)\in\mathbb{R}^{d}$ and log-variance $\log\sigma^2(x)\in\mathbb{R}^{d}$ of a diagonal Gaussian in $Z$ (latent dimension $d{=}20$).
We sample $z=\mu+\sigma\odot\epsilon,\ \epsilon\sim N(0,I)$.
The decoder $p_\theta(x\!\mid\!z)$ maps $z$ through a linear layer back to the flattened conv feature size, then uses two stride-2 transposed convolutions (64$\to$32, 32$\to C$; kernel $4{\times}4$, ReLU) with a final sigmoid to produce per-cell, per-object probabilities $\hat{x}\in[0,1]^{C\times H\times W}$.




\bibliographystyle{IEEEtran}
\bibliography{main}

\end{document}